\pdfoutput=1

\documentclass[11pt]{article}

\usepackage{ACL2023}
\usepackage{annotates}

\usepackage{times}
\usepackage{latexsym}

\usepackage[T1]{fontenc}

\usepackage[utf8]{inputenc}

\usepackage{microtype}

\usepackage{inconsolata}
\usepackage{booktabs}
\usepackage{hyperref}
\usepackage{booktabs} 
\usepackage{CJKutf8}
\usepackage{tabularx}
\usepackage{array}
\usepackage{CJKutf8}
\usepackage{adjustbox}
\usepackage{multirow}
\usepackage{colortbl}
\usepackage{graphicx}
\usepackage[bottom]{footmisc}
\usepackage{lipsum}

\title{Pragmatic Competence Evaluation of Large \\Language Models for the Korean Language}

\author{Dojun Park$^{1}$ \hspace{10pt} Jiwoo Lee$^{1, 2}$ \hspace{10pt} Hyeyun Jeong$^{1, 3, 4}$ \hspace{10pt} Seohyun Park$^{1, 4}$ \hspace{10pt} Sungeun Lee$^{1, 2, 4}$ \\
$^1$AI Institute of Seoul National University (AIIS) \\
$^2$Department of German Language and Literature, Seoul National University \\
$^3$Department of Korean Language and Literature, Seoul National University \\
$^4$Brain Humanities Lab (BHL), Seoul National University \\
\texttt{\{dojun.parkk, seohyun.parkk88\}@gmail.com} \\
\texttt{\{lee9055, tosirihy, cristlo5\}@snu.ac.kr}}

\begin{document}

\maketitle

\begin{abstract}

Benchmarks play a significant role in the current evaluation of Large Language Models (LLMs), yet they often overlook the models' abilities to capture the nuances of human language, primarily focusing on evaluating embedded knowledge and technical skills. To address this gap, our study evaluates how well LLMs understand context-dependent expressions from a pragmatic standpoint, specifically in Korean. We use both Multiple-Choice Questions (MCQs) for automatic evaluation and Open-Ended Questions (OEQs) assessed by human experts. Our results show that GPT-4 leads with scores of 81.11 in MCQs and 85.69 in OEQs, closely followed by HyperCLOVA X. Additionally, while few-shot learning generally improves performance, Chain-of-Thought (CoT) prompting tends to encourage literal interpretations, which may limit effective pragmatic inference. Our findings highlight the need for LLMs to better understand and generate language that reflects human communicative norms. The test set is publicly available on our GitHub repository at \url{https://github.com/DojunPark/pragmatic_eval_korean}.


\end{abstract}

\section{Introduction}

Research on LLMs has seen rapid advancements in recent years \cite{llm2, llm1}. Notably, ChatGPT was released in November 2022 and exemplifies this remarkable technological advancement. It has demonstrated impressive capabilities in a broad spectrum of Natural Language Processing (NLP) tasks, ranging from traditional ones like sentiment analysis and translation \cite{easynlp1, easynlp2} to more demanding areas such as complex problem-solving and creative writing \cite{diffnlp1, diffnlp2}. The versatility of ChatGPT has not only drawn significant attention from NLP researchers but also captivated the general public and tech companies, prompting many to develop their own LLMs \cite{bard, llama}.

The evaluation of LLMs is as crucial as their development. It enables the measurement of their performance for the targeted tasks and ensures that these models align with the anticipated standards of capability \cite{llmeval}. Systematic evaluation uncovers both the strengths and weaknesses of LLMs, which makes further fine-tuning of the models possible. This cycle of development and evaluation is essential in advancing these models, contributing to the creation of more reliable and suitable LLMs across diverse domains of real-world applications \cite{strongweak, domain}.

Benchmarks are serving a significant role in the current evaluation of LLMs \cite{openllm}. They provide task-specific datasets aligned with metrics, creating standardized scenarios for consistent evaluation. The primary advantage of these benchmarks is automating the evaluation process and enabling fair comparisons of models trained by different developers using varied strategies. However, the current benchmarking approach has notable limitations: a predominant focus on aspects like reasoning, computation, and knowledge \citep{arc, math, mmlu} with an emphasis on literal meaning, rather than implied meanings that vary with context, which are crucial for human-like language understanding. Additionally, many benchmarks rely on MCQs, a format that, while convenient for automated evaluation, does not fully evaluate the generative capacities of LLMs \cite{mcqlimit}. Furthermore, current benchmarks are showing a clear tendency to English-centric evaluation, which results in the under-exploration of LLMs' multilingual capacities \cite{llmsurvey, helm}.

\begin{table}[t]
\centering
\small
\begin{tabular}{@{}ll@{}}
\toprule
Statement         & \textit{"There's pizza in the fridge.}"                                                   \\ \midrule
Literal meaning   & Pizza is present inside the fridge.\\
Implicated meaning 1 & You are allowed to eat the pizza.                                              \\
Implicated meaning 2 & I won't cook dinner for you.                                                   \\ \bottomrule
\end{tabular}
\caption{Variations in Pragmatic Interpretation Based on Context}
\label{tab:pragmatic_interpretation}
\end{table}

Pragmatics is a linguistic study dealing with understanding language beyond just the literal meanings of words. It involves interpreting both the explicit (literal) and implicit (nonliteral) aspects of language, heavily influenced by context \cite{grice}. As an illustration, consider Table \ref{tab:pragmatic_interpretation}, which presents a statement with multiple possible interpretations. The literal interpretation is straightforward: pizza is inside the fridge. However, the implicated meanings vary with context. In the first scenario, e.g., imagine a friend visiting and expressing hunger; the statement might imply permission to eat the pizza. In the second scenario, e.g., consider a couple recovering from an argument. Here, the same statement might carry an undertone of reluctance to cook, reflecting the strained atmosphere. These examples illustrate that the pragmatic interpretation of a simple statement can vary significantly, transforming it into complex, context-dependent communication.

While earlier NLP models primarily focused on syntactic and semantic aspects of human language, with little emphasis on pragmatics, current LLMs necessitate a more comprehensive evaluation that extends beyond these traditional aspects \cite{nlpprag2, nlpprag1}. The enhanced performance of these models across diverse NLP tasks marks the significant need for evaluating their contextual language comprehension. Pragmatic understanding is especially crucial for conversational setups where AI assistants are expected to understand and respond in ways that meet human communicative needs \cite{prag4ai2, prag4ai}.

In this paper, we demonstrate a systematic evaluation of LLMs' pragmatic competence for the Korean language by analyzing it through four Gricean maxims: quantity, quality, relation, and manner, which are essential for understanding conversational implicature. Through this analysis, we aim to narrow the gap between the rapidly evolving capabilities of LLMs and the nuanced, human-level evaluation of language, ultimately suggesting directions for enhancing AI systems' awareness of contextual nuances.

Our study provides three main contributions:

\begin{itemize}
    \item We introduce the first dedicated resource for pragmatic evaluation of LLMs in Korean.
    \item We conduct a comprehensive evaluation of LLMs through MCQ and OEQ setups, assessing both automatic and qualitative dimensions of text generation.
    \item We explore the effectiveness of in-context learning strategies, specifically few-shot learning \cite{fewshot} and CoT reasoning \cite{cot}, to demonstrate their potential in enhancing LLM performance.
\end{itemize}

\section{Related Work}

\begin{table}[t]
\small
\centering
\begin{tabular}{@{}cp{6cm}@{}}
\toprule
\textbf{Maxim} & \textbf{Description}\\ \midrule
Quantity & Make your contribution as informative as is required.\\
Quality & Try to make your contribution one that is true.\\
Relation & Ensure that all the information you provide is relevant to the current conversation.\\
Manner & Be perspicuous; Be brief and orderly, and avoid obscurity and ambiguity.\\ \bottomrule
\end{tabular}
\caption{Gricean Maxims of Conversational Implicature}
\label{tab:maxims}
\end{table}

\subsection{Gricean Conversational Maxims}

In pragmatics, implicature refers to the meanings that speakers imply but do not explicitly state, which listeners must deduce from contextual cues. This aspect is essential for effective communication as humans often rely on implied meanings rather than explicit statements in real-world conversations. 

Grice outlined the cooperative principle as the foundation for rational conversation. This principle states that participants should make contributions that are appropriate for the current stage of the conversation, guided by its accepted purpose or direction \cite{grice}. This principle is further categorized into four conversational maxims, as demonstrated in Table \ref{tab:maxims}, which are crucial for understanding implicated meanings in communication. Conversational implicatures are often expressed by intentionally flouting these maxims, which leads to implicated meanings beyond the literal.

\subsection{Evaluating Pragmatic Competence of LLMs}
There have been efforts to assess the pragmatic capabilities of LLMs. \citet{pietro} assessed ChatGPT's pragmatic skills in Italian with the APACS Test \cite{apacs}, focusing on categories such as figurative language, humor, and interviews. Their results show that although ChatGPT closely mirrors human pragmatic understanding, it tends to be overly informative and struggles with text-based inferences, physical metaphors, and humor comprehension. However, The lack of transparency regarding the full test set limits how their findings align with further research.

\citet{bojic} evaluated LLMs against Grice's Cooperative Principle and its four maxims, reporting that the LLMs' performance exceeded the human average, with GPT-4 scoring the highest. This study, however, was limited by a participant pool of non-native English speakers and a small number of test items (twenty total). These factors may not accurately reflect native English speakers' pragmatic competence, suggesting a need for a larger, publicly available test set for more reliable evaluations of LLMs.

\subsection{Korean-Specific LLMs and its Evaluation}

The development of Korean-specific open-source LLMs has been accelerated by the introduction of the Open Ko-LLM Leaderboard \cite{open-ko-llm}, which features five benchmarks. KMMLU \cite{kmmlu} also emerges as an important benchmark, specifically designed to evaluate LLMs' capabilities in Korean across 45 diverse categories.

While not targeting an LLM, Nam et al. \cite{nam} evaluated the pragmatic competence of an AI speaker in Korean, using Gricean maxims in a multi-turn dialogue setup. They found that the maxim of relation was the most frequently violated by the AI speaker. Despite these efforts, research into the pragmatic abilities of LLMs for Korean is still in its early stages, underscoring the need for more specialized studies of these LLMs' pragmatic understanding.

\begin{CJK}{UTF8}{}
\CJKfamily{nanummj}
\begin{table*}[t]
\small
\centering
\begin{tabular}{@{}p{0.08\textwidth}p{0.42\textwidth}p{0.44\textwidth}@{}}
\toprule
 & \textbf{Korean} & \textbf{English Translation} \\
\midrule
\begin{tabular}[c]{@{}l@{}}\textbf{Context \&} \\ \textbf{Statement} \end{tabular} & 
\begin{tabular}[c]{@{}l@{}}철수가 바이올린을 연습하자 옆에서 듣던 영희가 말\\했다.\\ \textit{"우리집 강아지가 더 잘 한다."} \end{tabular} &
\begin{tabular}[c]{@{}l@{}}As Cheol-su was practicing the violin, Yeong-hee, who \\was listening next to him, said, \\ \textit{"My dog plays it better."} \\ \end{tabular} \\
\midrule
\begin{tabular}[c]{@{}l@{}}\textbf{MCQ} \end{tabular} &
\begin{tabular}[c]{@{}l@{}}다음 보기에서 위 발화가 갖는 가장 적절한 의미를 \\고르세요. \\ \textbf{(1) 철수의 바이올린 연주가 형편없다.} \\ (2) 영희가 기르는 강아지는 철수보다 바이올린을 잘 \\\hspace{0.5cm}켠다. \\ (3) 철수의 바이올린 연주는 강아지도 감동시킬 만큼 \\\hspace{0.5cm}훌륭하다. \\ (4) 철수는 고양이를 키우고 있다.\end{tabular} &
\begin{tabular}[c]{@{}l@{}}Choose the most appropriate meaning of the statement \\above from the \\options below. \\ \textbf{(1) Cheol-su's violin performance is terrible.} \\ (2) The dog raised by Yeong-hee plays the violin better \\\hspace{0.5cm}than Cheol-su. \\ (3) Cheol-su's violin performance is so excellent that \\\hspace{0.5cm}it can even move a dog. \\ (4) Cheol-su is raising a cat.\end{tabular} \\
\midrule
\textbf{OEQ} &
위 발화가 갖는 가장 적절한 의미를 서술하세요. &
Describe the most appropriate meaning of the statement above. \\
\bottomrule
\end{tabular}
\caption{Example of a Test Unit on the Maxim of Quality. The answer (1) in bold is the correct answer, as it accurately conveys the implicated meaning of the statement within the provided context.}
\label{tab:test_set_example}
\end{table*}

\end{CJK}

\section{Methodology}

\subsection{Constructing Pragmatic Test Set}

The development of the pragmatic test set was planned to thoroughly assess the nuanced understanding of conversational implicatures by LLMs. Below are the detailed considerations involved in the test set construction:

\begin{itemize}
    \item \textbf{Selection of Maxims}: We chose Grice’s maxims as the foundational framework due to their comprehensive coverage of conversational implicatures. These maxims are essential for assessing a model's ability to interpret meanings beyond literal words.
    
    \item \textbf{Test Set Size and Distribution}: The test set comprises 120 units, with 30 units allocated to each of the four maxims. This distribution ensures a balanced assessment across different aspects of pragmatic competence while allowing for statistically significant results.
    \item \textbf{Contextual Design}: Each test unit consists of a context that sets the scene for the dialogue, a statement made by one of the dialogue participants, and a follow-up question that asks the expressed meaning of the statement.
    \item \textbf{Expert Collaboration}: The test units were crafted by four experts holding master's degrees in linguistics or related fields, ensuring high-quality and contextually rich scenarios.
\end{itemize}


Table \ref{tab:test_set_example} presents an example from our test set, demonstrating the case of the maxim of quality. The example is shown in both Korean and its English translation. In this instance, the statement by Yeong-hee, \textit{``My dog plays it better,''} may seem to simply praise the dog's abilities if taken at face value. However, within the provided context of Cheol-su practicing the violin, it implies a critical judgment, suggesting that Cheol-su's violin playing is exceptionally poor.



This test set is designed to challenge LLMs across diverse conversational scenarios, assessing their ability to interpret implicature in a manner akin to human understanding. The inherent difficulty of each maxim varies, with the maxim of manner often introducing ambiguity that poses additional challenges \cite{Hoffmann}. We aim to empirically quantify their level of pragmatic competence, offering insights into their capabilities and limitations in processing human language.

\subsection{Multiple-Choice Questions vs. Open-ended Questions}

Table \ref{tab:test_set_example} outlines two types of evaluation questions: MCQs and OEQs. MCQs test a model's ability to select the most appropriate meaning from provided options, suitable for automated scoring but limited in assessing deeper generative and inferential skills. In contrast, OEQs demand a narrative response, enabling experts to judge the depth and context appropriateness of the answers. This dual strategy evaluates both the basic comprehension and the more complex generative abilities of LLMs.

For MCQs, each question is accompanied by four options. We categorized these options to represent distinct types of interpretation: the correct answer that accurately reflects the implicated meaning within the context, a naive literal interpretation, an incorrect interpretation within context, and an incorrect interpretation out of context. A response is considered correct if the LLMs' generated answer explicitly selects the option number corresponding to the correct interpretation.

For OEQs, three independent assessors with qualifications matching those of the test set creators evaluate LLMs' narrative responses using a Likert scale from 1 to 5. A score of 5 denotes perfect contextual understanding and accurate interpretation of implicature, whereas a score of 1 indicates a complete misunderstanding of both context and literal meaning. Scores are subsequently re-scaled to a 0-100 range for comparison with MCQ outcomes.

\subsection{In-Context Learning}

Recent research shows that in-context learning allows LLMs to quickly adapt to new tasks using their pre-existing knowledge base, without prior training \cite{icl1, icl2}. This study examines two specific strategies: few-shot learning \cite{fewshot} and CoT prompting \cite{cot}, as detailed in Table \ref{tab:in-context-setup}. We use the MCQ format to compare the impact of these strategies on LLMs' pragmatic competence across six different setups.

We define three few-shot learning scenarios based on example quantity: zero-shot, one-shot (one example illustrating the maxim of quality), and four-shot (four examples each demonstrating a different maxim). We also compare two CoT prompting setups: the base setup, which only presents test questions and answers, and the CoT setup, which includes detailed reasoning for each question, prompting LLMs to articulate their inferential processes as demonstrated in the few-shot examples.

\begin{CJK}{UTF8}{}
\CJKfamily{nanummj}
Specifically, for the zero-shot scenario with CoT, we adopt the methodology of \citet{zeroshot}, by appending the phrase “답: 순차적으로 생각해봅시다.” (translated as “Answer: Let’s think step by step.”) at the end of the question, guiding the model towards a structured inferential approach.
\end{CJK}

\begin{table}[t]
\centering
\small
\begin{tabular}{@{}ccc@{}}
\toprule
\textbf{Setup} & \textbf{Few-Shot Examples} & \textbf{CoT Prompting} \\ \midrule
0-shot (Base) & 0 & X \\
1-shot (Base) & 1 & X \\
4-shots (Base) & 4 & X \\
0-shot (CoT) & 0 & O \\
1-shot (CoT) & 1 & O \\
4-shots (CoT) & 4 & O \\ \bottomrule
\end{tabular}
\caption{Experimental Setups for Assessing LLMs' In-Context Learning Capabilities in the MCQ Test}
\label{tab:in-context-setup}
\end{table}

\section{Experiment}

\begin{table*}[t]
\centering
\begin{tabular}{@{}lrrrrr@{}}
\toprule
 & \multicolumn{1}{l}{\textbf{Quantity}} & \multicolumn{1}{l}{\textbf{Quality}} & \multicolumn{1}{l}{\textbf{Relation}} & \multicolumn{1}{l}{\textbf{Manner}} & \multicolumn{1}{l}{\textbf{Avg.}} \\ \midrule
GPT-3.5 & 36.67 & 50.00 & 28.89 & 44.44 & \textbf{40.00} \\
GPT-4 & {\underline{82.22}} & 90.00 & {\underline{82.22}} & {\underline{70.00}} & \textbf{{\underline{81.11}}} \\
Gemini-Pro & 62.22 & 75.56 & 53.33 & 47.78 & \textbf{59.72} \\
HyperClova X & 67.78 & {\underline{93.33}} & 47.78 & 61.11 & \textbf{67.50} \\
LDCC-Solar & 57.78 & 57.78 & 36.67 & 45.56 & \textbf{49.44} \\ \bottomrule
\end{tabular}
\caption{Scores on MCQ Test Across four Gricean Maxims}
\label{tab:mcq_scores}
\end{table*}

\begin{figure*}[t]
\centering
\includegraphics[width=0.7\textwidth]{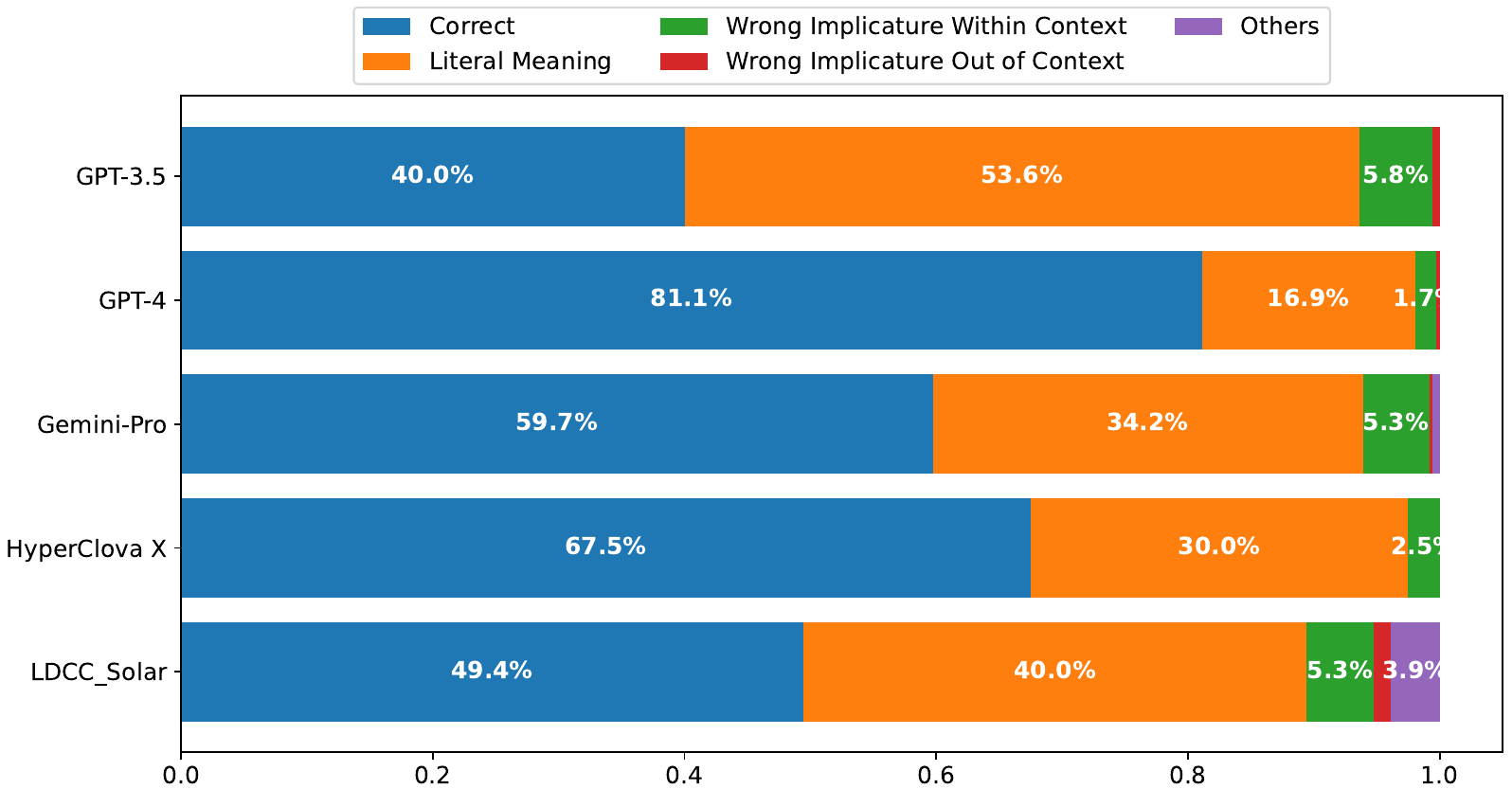}
\caption{MCQ Option Choice Distribution by each LLM}
\label{fig:mcq_dist}
\end{figure*}

\subsection{Experimental Setup}

\paragraph{Model Selection.} Our study compares five LLMs: GPT-3.5-turbo and GPT-4 \cite{gpt-4} by OpenAI, Gemini-Pro \cite{gemini} by Google DeepMind, HyperCLOVA X \cite{clova} by NAVER, and LDCC-Solar \cite{ldcc} by Lotte Data Communication. GPT-3.5, GPT-4, and Gemini-Pro are multilingual, capable of processing Korean among other languages. HyperCLOVA X and LDCC-Solar, however, are optimized specifically for Korean.

\paragraph{Hyperparameter Setting.} For each LLM, we uniformly configured the hyperparameter settings to ensure a fair comparison across the board. The maximum output length was set to 512 tokens, and the temperature parameter was set to 0.7. For HyperCLOVA X, which was not accessible via API, we manually retrieved each response from its website and initiated a new session for each interaction to maintain consistency with the other LLMs.

\paragraph{Performance Report.} We generated three responses from each LLM for each test unit to report their performance. In the case of the MCQ tests, we varied the order of the provided options to mitigate potential bias towards any specific answer position. We report the averaged scores from three trials to ensure the reliability of our results, considering the variability in the LLMs' outputs.

\begin{table*}[t]
\centering
\begin{tabular}{@{}lrrrrr@{}}
\toprule
 & \multicolumn{1}{l}{\textbf{Quantity}} & \multicolumn{1}{l}{\textbf{Quality}} & \multicolumn{1}{l}{\textbf{Relation}} & \multicolumn{1}{l}{\textbf{Manner}} & \multicolumn{1}{l}{\textbf{Avg.}} \\ \midrule
GPT-3.5 & 61.25 & 49.75 & 67.50 & 64.75 & \textbf{60.81} \\
GPT-4 & {\underline{82.25}} & {\underline{88.25}} & {\underline{94.25}} & {\underline{78.00}} & {\underline{\textbf{85.69}}} \\
Gemini-Pro & 75.00 & 64.75 & 77.00 & 68.50 & \textbf{71.31} \\
HyperClova X & 81.50 & 83.50 & 88.00 & 73.25 & \textbf{81.56} \\
LDCC-Solar & 62.25 & 54.00 & 62.50 & 49.25 & \textbf{57.00} \\ \bottomrule
\end{tabular}
\caption{Scores on OEQ Test Across four Gricean Maxims}
\label{tab:oeq_scores}
\end{table*}

\begin{figure*}[t]
\centering
\includegraphics[width=0.65\textwidth]{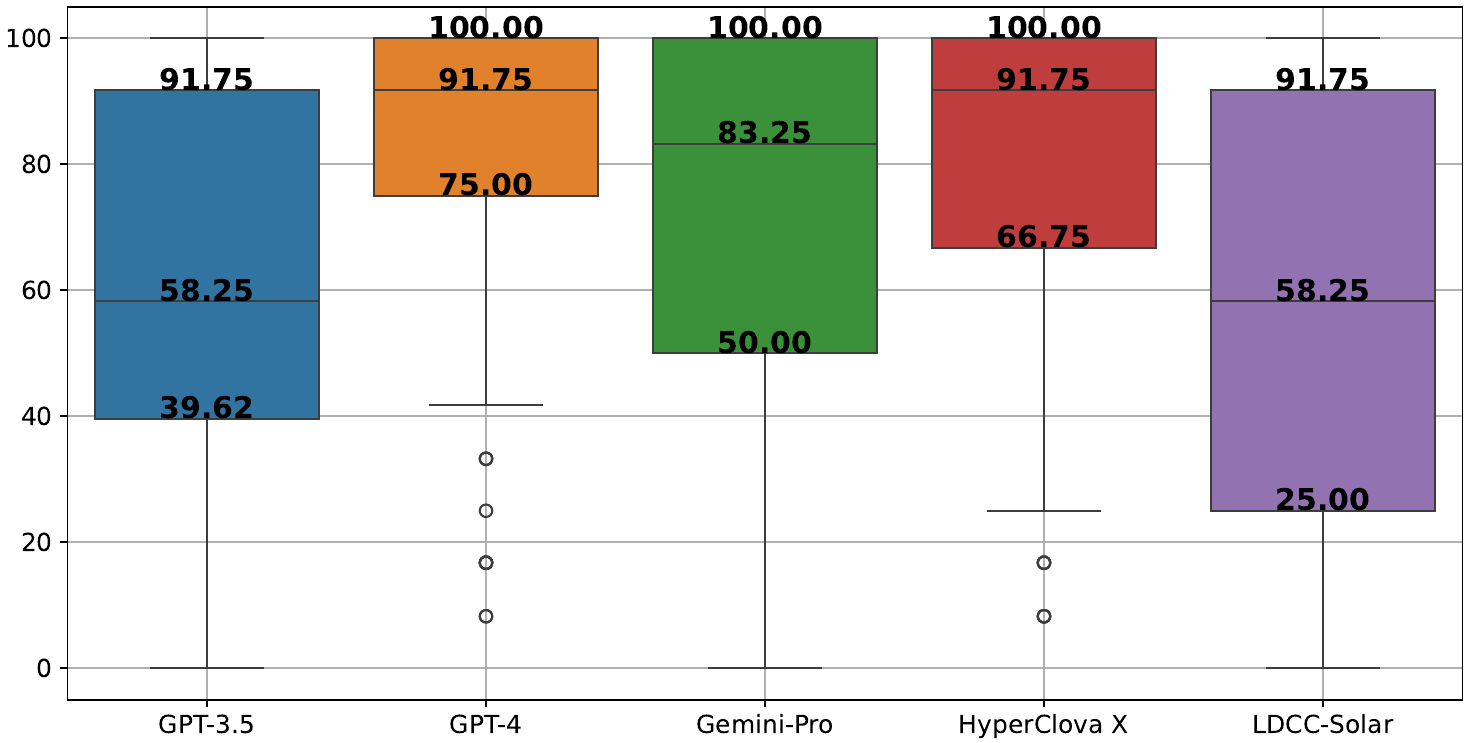}
\caption{Box Plots of OEQ Score Distributions by each LLM}
\label{fig:oeq_dist}
\end{figure*}

\subsection{Result of the MCQ Test}

\paragraph{Analysis of LLM Performance.}
Table \ref{tab:mcq_scores} illustrates the performance of each LLM on the MCQ test across the four Gricean maxims, along with the overall average score. GPT-4 leads with an impressive average score of 81.11, significantly outperforming all compared LLMs. HyperCLOVA X and Gemini-Pro follow closely with scores of 67.5 and 59.72, respectively, showcasing their proficiency in understanding conversational implicature.

Notably, LDCC-Solar, with nearly half the parameter size of GPT-3.5-turbo--10.7 billion compared to the reported 20 billion \cite{parameters}-- manages to exceed GPT-3.5's performance by 9.44 points. This highlights the effectiveness of LLMs specialized for Korean and suggests that a larger parameter size does not necessarily guarantee better performance.

Conversely, the significant performance gap between GPT-3.5 and GPT-4, which boasts 1.7 trillion parameters, underscores the continued importance of parameter scale. This difference emphasizes that while specialized training is crucial, the scale of parameters remains a critical factor in facilitating a model's capabilities, particularly in tasks requiring nuanced pragmatic inference.

In the evaluation of individual maxims, LLMs consistently score higher on the maxim of quality, while they tend to receive lower scores for the maxims of relation and manner. This pattern suggests that, within the MCQ setup, the presence of options based on literal interpretations may inadvertently make it more challenging to choose answers that align with the appropriate implicature, particularly for the maxims of relation and manner. The reason behind this is twofold: For the maxim of quality, literal interpretations often lack semantic plausibility, making them easier for LLMs to rule out. In contrast, for the contexts governed by relation and manner, options with literal meanings can be often semantically valid, complicating the selection process.

Notably, HyperCLOVA X's superior performance in the maxim of quality, surpassing even GPT-4, can likely be attributed to its language-specific training. This implies the advantage of developing a model with a comprehensive dataset in Korean, which enables it to achieve a deeper understanding of intricate linguistic nuances.

\paragraph{LLMs' Answer Selection Examined.}
\label{sec:mcq_choice}

Figure \ref{fig:mcq_dist} demonstrates the distribution of option types selected by the LLMs during the MCQ test, revealing a noticeable preference across all models for literal interpretations among wrong answer options. This tendency underscores the challenge posed by a bias towards literal meaning, which obstructs effective pragmatic inference. Specifically, GPT-3.5 shows a pronounced preference for literal interpretations, selecting them 53.6\% of the time, compared to only 40\% for correct pragmatic interpretations.

The analysis further reveals that, even when selecting incorrect answers, all LLMs are more inclined to choose options that are incorrect within the given context rather than options that are out of context. This behavior suggests that the models do engage with the context in their decision-making, albeit not always successfully leading to the correct implicature.

LDCC-Solar, in particular, shows a tendency to select options not provided in the prompt—such as `5' or `d', which it generates on its own, at a rate of 3.9\%—underscoring a significant challenge in adhering to the given choices and further deviating from accurate pragmatic inference. Additionally, LDCC-Solar's responses often include noisy text, irrelevant material that detracts from the response quality, pointing to challenges that go beyond understanding pragmatic cues.

\subsection{Result of the OEQ Test}

\paragraph{Performance Evaluation from MCQs to OEQs.}

In Table \ref{tab:oeq_scores} showcasing the results of the OEQ test, GPT-4 maintains its leading performance, with HyperCLOVA X and Gemini-Pro closely following, consistent with the MCQ test findings. However, while the performance gap between GPT-4 and HyperCLOVA X was 13.61 in the MCQ test, it has significantly narrowed to 4.13 in the OEQ test, indicating HyperCLOVA X's enhanced capability in the narrative setup. 

Conversely, LDCC-Solar, which outperformed GPT-3.5 in the MCQ test by 9.44, falls behind GPT-3.5 by 3.81 in the OEQ test, suggesting that while LDCC-Solar excels in option selection, GPT-3.5 demonstrates superior generative abilities. 

These discrepancies underscore a critical consideration for LLM development, highlighting that the MCQ framework, while popular for benchmarking, may not fully capture the essence of LLMs' generative capabilities. Given that LLMs often operate in conversational setups without predefined options in real-world scenarios, the significance of qualitative evaluation in assessing LLM narrative generation capabilities becomes evident.

For all maxims, GPT-4 consistently achieves the highest scores, closely followed by HyperCLOVA X. Similar to the MCQ test, the maxim of manner yields lower scores overall, yet notably, the maxim of relation receives the highest scores in the OEQ test, with the maxim of quality scoring relatively lower, which are contrastive to the MCQ results. This shift likely suggests that the narrative format of OEQs, devoid of predefined options, simplifies the task of adhering to generating correct answers at the maxim of relation for LLMs, reducing its complexity. Conversely, this format appears to diminish the advantages previously observed for the maxim of quality in the MCQ setup.

Interestingly, LDCC-Solar exhibits a unique behavior by initially generating options akin to those in the MCQ format and then selecting one, even when asked to respond narratively. This behavior likely results from overfitting to the MCQ format, which predominates the benchmarks used in the Open Ko-LLM Leaderboard. Such a strategy, while innovative, may not always align with the expectations for narrative answers and could reflect a limitation in the model's adaptability to varied question formats. Additionally, as observed in the MCQ test, LDCC-Solar's responses often include noisy text that diminishes the overall quality of its responses. 

\begin{table*}[t]
\centering
\begin{tabular}{@{}lrrrrrr@{}}
\toprule
 & \multicolumn{2}{c}{\textbf{0-shot}} & \multicolumn{2}{c}{\textbf{1-shot}} & \multicolumn{2}{c}{\textbf{4-shots}} \\ \cmidrule(l){2-3} \cmidrule(l){4-5} \cmidrule(l){6-7} & \multicolumn{1}{c}{base} & \multicolumn{1}{c}{CoT} & \multicolumn{1}{c}{base} & \multicolumn{1}{c}{CoT} & \multicolumn{1}{c}{base} & \multicolumn{1}{c}{CoT} \\ \midrule
GPT-3.5 & \textbf{40.00} & 35.56 & \textbf{44.72} & 40.56 & \textbf{56.94} & 47.50 \\
GPT-4 & \textbf{{\underline{81.11}}} & {\underline{77.22}} & \textbf{{\underline{86.11}}} & {\underline{80.56}} & \textbf{{\underline{89.17}}} & {\underline{84.44}} \\
Gemini-Pro & \textbf{59.72} & 59.17 & \textbf{63.06} & 61.94 & \textbf{66.39} & 62.78 \\
HyperClova X & \textbf{67.50} & 66.67 & \textbf{78.06} & 62.50 & \textbf{84.14} & 71.94 \\
LDCC-Solar & 49.44 & \textbf{52.50} & \textbf{59.44} & 0.00 & 0.00 & 0.00 \\ \bottomrule
\end{tabular}
\caption{Scores on MCQ Test Under Few-Shot Learning Conditions with and without Chain of Thought prompting}
\label{tab:in-context-results}
\end{table*}

\paragraph{Analyzing Score Distributions of LLMs.}

Figure \ref{fig:oeq_dist} presents box plots that illustrate the score distributions of each LLM in the OEQ test. In these plots, GPT-4 and HyperCLOVA X exhibit the same median (Q2) and 75th percentile (Q3) values, indicating comparable performance at the median and upper quartiles. However, at the 25th percentile (Q1), representing the lowest 25\% of scores, GPT-4 demonstrates superior performance with a score of 75 compared to HyperCLOVA X's 66.75. This suggests that GPT-4 maintains a higher level of quality for the lower-performing test units, indicating greater overall stability in its responses.

Gemini-Pro displays a broader interquartile range between Q1 and Q3, indicating less consistency in its performance compared to GPT-4 and HyperCLOVA X. This wider range suggests variability in the quality of Gemini-Pro's responses across different test units.

Similarly, GPT-3.5 and LDCC-Solar show no differences in their Q2 and Q3 values. However, the Q1 score for LDCC-Solar is markedly lower at 25 compared to 39.62 for GPT-3.5. This highlights that the lower-quality responses from LDCC-Solar contribute significantly to the reduced consistency in response quality.

\subsection{Results of In-Context Learning}

Table \ref{tab:in-context-results} showcases the impact of employing two in-context learning techniques in the MCQ setup: few-shot learning and CoT prompting. Consistent with our findings from the MCQ test, GPT-4 continues to outperform all other LLMs across different setups, with HyperCLOVA X and Gemini-Pro following closely. The application of the few-shot learning strategy leads to incremental improvements in performance for most LLMs in both setups with and without CoT prompting.

Notably, HyperCLOVA X exhibits a remarkable improvement by few-shot learning in the base setup, achieving a higher increase compared to other LLMs; it gains 10.56 points moving from 0-shot to 1-shot learning, and an additional 6.08 points transitioning from 1-shot to 4-shots, resulting in a score of 84.14, which approaches GPT-4’s leading 89.17. Such impressive improvements indicate HyperCLOVA X’s underlying capacities in pragmatic competence and highlight its effectiveness as a Korean-specific LLM, which has evidently benefited from comprehensive pre-training on a diverse Korean text corpus.

In contrast to the advantages observed with the few-shot technique, applying CoT prompting appears to diminish the performance of LLMs compared to their base setups without CoT. Notably, the CoT approach tends to introduce a bias towards literal interpretations. This tendency is particularly evident in the 4-shot setup, where LLMs frequently select multiple options, often including those of literal interpretations.

The context units provided in the evaluation data set were limited to 1-2 sentences, with information crucial for pragmatic inferences not explicitly delineated at the semantic level. In such cases, the CoT method arguably obstructs the inherent capabilities of LLMs for pragmatic inference. This highlights a limitation of CoT, especially in scenarios where pragmatic cues are subtly embedded below the semantic level and require nuanced interpretation. While CoT has demonstrated potential in enhancing logical thinking and problem-solving in contexts with explicit statements, its effectiveness for pragmatic inference appears contingent on the depth and type of contextual information provided.

\begin{CJK}{UTF8}{}
\CJKfamily{nanummj}
Differently from other LLMs, LDCC-Solar exhibits a slight increase in score for the 0-shot CoT setup, indicating some initial advantages of this strategy. However, its performance drastically declines in the 1-shot with CoT and both 4-shot setups, with and without CoT, where it completely fails to generate the required answers, resulting in scores of 0. This failure appears to stem from LDCC-Solar's tendency to reiterate the given prompt in its responses, which suggests a limitation in its processing capabilities when faced with increased complexity or specificity in the tasks. In most responses, it either repeats the prompts without adding any substantive content or produces irrelevant single words, such as '철수' (Cheol-su, a person’s name used in prompts), or meaningless fragments like '제' or '먼'. These responses highlight the model's difficulty in moving beyond a straightforward single-shot example, possibly reflecting the limits of its capacity to handle nuanced or layered instructions.

\begin{table*}[t]
\centering
\small
\begin{tabular}{@{} >{\centering\arraybackslash}m{0.5cm} >{\centering\arraybackslash}m{1cm} >{\raggedright\arraybackslash}p{6.5cm} >{\raggedright\arraybackslash}p{6.8cm} @{}} 
\toprule
\textbf{Id} & \textbf{Maxim} & \multicolumn{1}{c}{
\begin{tabular}[c]{@{}c@{}}\textbf{Context and Statement in Korean} \end{tabular}} 
& \multicolumn{1}{c}{
\begin{tabular}[c]{@{}c@{}}\textbf{English Translation} \end{tabular}} 
\\ \midrule
27 & Quantity & \begin{tabular}[c]{@{}l@{}}
붕어빵 가게 앞을 지나가던 철수가 영희에게 \\
현금이 있는지 물었고 영희는 다음과 같이 말했다. \\ 
\textit{‘5만원짜리 밖에 없어’}.\end{tabular} & \begin{tabular}[c]{@{}l@{}}
As Chul-su was passing by a 붕어빵(bung-eo-bbang) \\
street stall, he asked Yeong-hee if she had any cash.\\ Yeong-hee replied, \\ 
\textit{‘I only have a 50,000 won bill’}.\end{tabular} \\ \midrule
83 & Relation & \begin{tabular}[c]{@{}l@{}}
철수 집에 놀러 간 영희는 주방에 많은 귤이 쌓여 \\
있는 것을 보고 귤이 왜 이렇게 많은지 물었고 \\
철수는 다음과 같이 말했다. \\ 
\textit{‘우리 작은 아버지께서 제주도에 사셔'}.\end{tabular} & \begin{tabular}[c]{@{}l@{}}
When Yeong-hee visited Chul-su’s house, she saw \\
many tangerines piled up in the kitchen and asked \\
why there were so many. Chul-su replied, \\ 
\textit{‘My uncle lives on Jeju Island’.}\end{tabular} \\ \bottomrule
\end{tabular}
\caption{Examples of Korean Culture-Specific Test Units}
\label{tab:culture-specific}
\end{table*}

\section{Case Study: In-Depth Analysis of LLM Responses to OEQs}

\subsection{A Closer Look at Strengths and Weaknesses}

GPT-4, the top performer in our evaluations, excels by offering clear, definitive answers rather than multiple interpretations. This simplifies decision-making for users by eliminating the need to filter through various possible answers. Moreover, its avoidance of uncertain expressions such as ‘-으로 예상할 수 있다’ (it can be speculated that) or ‘-일 수 있다’ (may be)—common in GPT-3.5 and occasionally seen in Gemini and HyperCLOVA X—further contributes to the reliability of its answers.

HyperCLOVA X demonstrates its strengths by providing detailed explanations of its reasoning process, significantly facilitating a deeper comprehension for users. Conversely, Gemini-Pro reveals a critical limitation by occasionally offering brief responses without explanation. An example of this is returning a single-word answer, ‘반어법’ (irony), without offering a supplementary interpretation.

\subsection{Analyzing Error Patterns in LLM Responses}

The most prevalent error arises from struggling to interpret implicatures, despite understanding the context, followed by errors due to purely literal interpretations, especially concerning the maxim of manner. This contrasts with our MCQ answer choice analysis (cf. Section \ref{sec:mcq_choice}), where literal interpretation errors were more prevalent. This indicates that the MCQ format may have prompted LLMs to favor literal meanings, hindering accurate pragmatic inferences.

LLMs also struggle with interpreting synonyms or phonetically similar words. A notable example is the word ‘사기’ (‘sagi’), meaning both ‘to buy’ and ‘fraud.’ Despite the context indicating the purchasing meaning, only HyperCLOVA X correctly identified it, while the others misinterpreted it as fraud.

We also observed errors associated with formatting in the responses. For example, Gemini-Pro occasionally included markdown formatting symbols into its output directly, like **표현** (`expression'), with the intention of emphasizing the text. However, these markers appeared verbatim in the responses, resulting in inaccurately formatted and potentially confusing answers.

\subsection{Responses to Cultural-specific questions in Korean}

Through our examination, we have observed that certain test units are deeply rooted in Korean culture and significantly influence the responses of LLMs. To exemplify this, we have chosen two prototypical instances, detailed in Table \ref{tab:culture-specific}.
 
The first involves 붕어빵 ('bung-eo-bbang'), a popular Korean street food made with fish-shaped molds filled with sweet red bean paste, especially enjoyed in winter. Typically sold at cash-only street stalls, bung-eo-bbang costs around 1,000 won for three pieces. In the example, the statement \textit{‘I only have a 50,000 won bill’}  can be interpreted as an expression of her reluctance to buy bung-eo-bbang, not just due to the inconvenience it would cause the stall owner in providing change, but also because of the impracticality for Yeong-hee to manage the received smaller bills. While several LLMs, including GPT-4, struggled to accurately infer the implied meaning of Yeong-hee's statement, HyperCLOVA X demonstrated a correct understanding, showcasing its ability to adjust responses based on the specific cultural context prevalent in Korea.

In the context of the second example, Jeju Island is renowned for its regional specialty, the Jeju-Tangerine, and it is a common practice among residents to send Jeju-Tangerines as gifts. Considering this context, the statement \textit{'My uncle lives on Jeju Island'} can be implicitly understood to mean \textit{'We have a lot of tangerines because my uncle, who lives on Jeju Island, sent them to us.'} Both GPT-4 and HyperCLOVA X excelled at figuring out the meaning behind the statement. Notably, even the LDCC-Solar, which generally scored lower on average across the board, achieved a high score in this specific case.

These findings align with the results of Son et al. \cite{kmmlu}, where HyperCLOVA X demonstrated its superior performance for Korea-specific knowledge compared to other LLMs, including GPT-4. This observation underscores the critical importance of developing LLMs that are capable of comprehending culture-specific contexts, which is essential not only for accurate general knowledge but also for higher-level processes, such as pragmatic inference.

\end{CJK}

\section{Conclusion}

In this study, we address an under-explored aspect of LLM evaluation—the pragmatic evaluation of LLMs, with a specific focus on Korean. We developed a test set comprising 120 test units rooted in Gricean theory of conversational implicature to rigorously assess the pragmatic competence of LLMs.

Our findings indicate that GPT-4 outperforms all competitors in both MCQ and OEQ formats, followed closely by HyperCLOVA X and Gemini-Pro. HyperCLOVA X notably reduces the performance gap seen in MCQs when tested in OEQs. In contrast, LDCC-Solar, an open-source LLM, surpasses GPT-3.5 in MCQs but underperforms in OEQs, highlighting the impact of the question format. Additionally, while few-shot learning generally improved LLM performance, CoT prompting had a detrimental effect, likely due to its focus on literal rather than pragmatic interpretations.

\section*{Limitations}
While our study provides a comprehensive evaluation of LLMs’ pragmatic competence, we identify two primary areas for enhancement in our future work. Firstly, although the test set of 120 units—30 for each Gricean maxim—has yielded meaningful insights, this quantity remains modest in comparison to other benchmarks commonly utilized for LLM evaluation. Additionally, while focusing on Korean has unveiled significant findings, the multilingual capabilities of LLMs are yet to be fully explored. Viewing this study as a pilot, we aim to develop a more comprehensive and reliable multilingual evaluation framework for LLMs' pragmatic competence in our future work.

\section*{Acknowledgment}
This research was supported by Basic Science Research Program through the National Research Foundation of Korea (NRF) funded by the Ministry of Education (RS-2023-00274280).

\bibliography{anthology,custom}

\begin{thebibliography}{40}
\expandafter\ifx\csname natexlab\endcsname\relax\def\natexlab#1{#1}\fi

\bibitem[{Achiam et~al.(2023)Achiam, Adler, Agarwal, Ahmad, Akkaya, Aleman, Almeida, Altenschmidt, Altman, Anadkat et~al.}]{gpt-4}
Josh Achiam, Steven Adler, Sandhini Agarwal, Lama Ahmad, Ilge Akkaya, Florencia~Leoni Aleman, Diogo Almeida, Janko Altenschmidt, Sam Altman, Shyamal Anadkat, et~al. 2023.
\newblock \href {https://arxiv.org/abs/2303.08774} {Gpt-4 technical report}.
\newblock \emph{arXiv preprint arXiv:2303.08774}.

\bibitem[{Arcara and Bambini(2016)}]{apacs}
Giorgio Arcara and Valentina Bambini. 2016.
\newblock \href {https://www.frontiersin.org/journals/psychology/articles/10.3389/fpsyg.2016.00070/full} {A test for the assessment of pragmatic abilities and cognitive substrates (apacs): Normative data and psychometric properties}.
\newblock \emph{Frontiers in psychology}, 7:70.

\bibitem[{Bojic et~al.(2023)Bojic, Kovacevic, and Cabarkapa}]{bojic}
Ljubisa Bojic, Predrag Kovacevic, and Milan Cabarkapa. 2023.
\newblock \href {https://arxiv.org/abs/2312.09545} {Gpt-4 surpassing human performance in linguistic pragmatics}.
\newblock \emph{arXiv preprint arXiv:2312.09545}.

\bibitem[{Bommasani et~al.(2023)Bommasani, Liang, and Lee}]{helm}
Rishi Bommasani, Percy Liang, and Tony Lee. 2023.
\newblock \href {https://arxiv.org/abs/2211.09110} {Holistic evaluation of language models}.
\newblock \emph{Annals of the New York Academy of Sciences}, 1525(1):140--146.

\bibitem[{Brown(2020)}]{fewshot}
Tom~B Brown. 2020.
\newblock \href {https://arxiv.org/abs/2005.14165} {Language models are few-shot learners}.
\newblock \emph{arXiv preprint ArXiv:2005.14165}.

\bibitem[{Chang et~al.(2024)Chang, Wang, Wang, Wu, Yang, Zhu, Chen, Yi, Wang, Wang et~al.}]{llmeval}
Yupeng Chang, Xu~Wang, Jindong Wang, Yuan Wu, Linyi Yang, Kaijie Zhu, Hao Chen, Xiaoyuan Yi, Cunxiang Wang, Yidong Wang, et~al. 2024.
\newblock \href {https://dl.acm.org/doi/10.1145/3641289} {A survey on evaluation of large language models}.
\newblock \emph{ACM Transactions on Intelligent Systems and Technology}, 15(3):1--45.

\bibitem[{Clark et~al.(2018)Clark, Cowhey, Etzioni, Khot, Sabharwal, Schoenick, and Tafjord}]{arc}
Peter Clark, Isaac Cowhey, Oren Etzioni, Tushar Khot, Ashish Sabharwal, Carissa Schoenick, and Oyvind Tafjord. 2018.
\newblock \href {https://arxiv.org/abs/1803.05457} {Think you have solved question answering? try arc, the ai2 reasoning challenge}.
\newblock \emph{arXiv preprint arXiv:1803.05457}.

\bibitem[{di~San~Pietro et~al.(2023)di~San~Pietro, Frau, Mangiaterra, and Bambini}]{pietro}
Chiara~Barattieri di~San~Pietro, Federico Frau, Veronica Mangiaterra, and Valentina Bambini. 2023.
\newblock \href {https://osf.io/preprints/psyarxiv/ckghw} {The pragmatic profile of chatgpt: assessing the pragmatic skills of a conversational agent}.

\bibitem[{Dong et~al.(2022)Dong, Li, Dai, Zheng, Wu, Chang, Sun, Xu, and Sui}]{icl1}
Qingxiu Dong, Lei Li, Damai Dai, Ce~Zheng, Zhiyong Wu, Baobao Chang, Xu~Sun, Jingjing Xu, and Zhifang Sui. 2022.
\newblock \href {https://arxiv.org/abs/2301.00234} {A survey on in-context learning}.
\newblock \emph{arXiv preprint arXiv:2301.00234}.

\bibitem[{Fourrier et~al.(2024)Fourrier, Habib, Lozovskaya, Szafer, and Wolf}]{openllm}
Clémentine Fourrier, Nathan Habib, Alina Lozovskaya, Konrad Szafer, and Thomas Wolf. 2024.
\newblock Open llm leaderboard v2.
\newblock \url{https://huggingface.co/spaces/open-llm-leaderboard/open_llm_leaderboard}.

\bibitem[{Ge et~al.(2024)Ge, Hua, Mei, Tan, Xu, Li, Zhang et~al.}]{domain}
Yingqiang Ge, Wenyue Hua, Kai Mei, Juntao Tan, Shuyuan Xu, Zelong Li, Yongfeng Zhang, et~al. 2024.
\newblock \href {https://papers.nips.cc/paper_files/paper/2023/hash/1190733f217404edc8a7f4e15a57f301-Abstract-Datasets_and_Benchmarks.html} {Openagi: When llm meets domain experts}.
\newblock \emph{Advances in Neural Information Processing Systems}, 36.

\bibitem[{Grice(1975)}]{grice}
Herbert~Paul Grice. 1975.
\newblock \href {https://www.ucl.ac.uk/ls/studypacks/Grice-Logic.pdf} {Logic and conversation}.
\newblock \emph{Syntax and semantics}, 3:43--58.

\bibitem[{Guo et~al.(2023)Guo, Jin, Liu, Huang, Shi, Yu, Liu, Li, Xiong, Xiong et~al.}]{llmsurvey}
Zishan Guo, Renren Jin, Chuang Liu, Yufei Huang, Dan Shi, Linhao Yu, Yan Liu, Jiaxuan Li, Bojian Xiong, Deyi Xiong, et~al. 2023.
\newblock \href {https://arxiv.org/abs/2310.19736} {Evaluating large language models: A comprehensive survey}.
\newblock \emph{arXiv preprint arXiv:2310.19736}.

\bibitem[{Hendrycks et~al.(2020)Hendrycks, Burns, Basart, Zou, Mazeika, Song, and Steinhardt}]{mmlu}
Dan Hendrycks, Collin Burns, Steven Basart, Andy Zou, Mantas Mazeika, Dawn Song, and Jacob Steinhardt. 2020.
\newblock \href {https://arxiv.org/abs/2009.03300} {Measuring massive multitask language understanding}.
\newblock \emph{arXiv preprint arXiv:2009.03300}.

\bibitem[{Hendrycks et~al.()Hendrycks, Burns, Kadavath, Arora, Basart, Tang, Song, and Steinhardt}]{math}
Dan Hendrycks, Collin Burns, Saurav Kadavath, Akul Arora, Steven Basart, Eric Tang, Dawn Song, and Jacob Steinhardt.
\newblock \href {https://arxiv.org/abs/2103.03874} {Measuring mathematical problem solving with the math dataset}.
\newblock \emph{Sort}, 2(4):0--6.

\bibitem[{Hoffmann(2010)}]{Hoffmann}
Ludger Hoffmann. 2010.
\newblock \href {https://www.degruyter.com/document/doi/10.1515/9783110226300/html?lang=en} {\emph{Sprachwissenschaft: ein Reader}}.
\newblock de Gruyter.

\bibitem[{Jiao et~al.(2023)Jiao, Wang, Huang, Wang, and Tu}]{easynlp2}
Wenxiang Jiao, Wenxuan Wang, Jen-tse Huang, Xing Wang, and Zhaopeng Tu. 2023.
\newblock \href {https://arxiv.org/abs/2301.08745} {Is chatgpt a good translator? a preliminary study}.
\newblock \emph{arXiv preprint arXiv:2301.08745}, 1(10).

\bibitem[{Kabbara(2019)}]{nlpprag2}
Jad Kabbara. 2019.
\newblock \href {https://aclanthology.org/N19-3010/} {Computational investigations of pragmatic effects in natural language}.
\newblock In \emph{Proceedings of the 2019 Conference of the North American Chapter of the Association for Computational Linguistics: Student Research Workshop}, pages 71--76.

\bibitem[{Khatun and Brown(2024)}]{mcqlimit}
Aisha Khatun and Daniel~G Brown. 2024.
\newblock \href {https://arxiv.org/abs/2401.07955} {A study on large language models' limitations in multiple-choice question answering}.
\newblock \emph{arXiv preprint arXiv:2401.07955}.

\bibitem[{Kim(2024)}]{ldcc}
Wonchul Kim. 2024.
\newblock Ldcc-solar-10.7b.
\newblock \url{https://huggingface.co/LDCC/LDCC-SOLAR-10.7B}.

\bibitem[{Kojima et~al.(2022)Kojima, Gu, Reid, Matsuo, and Iwasawa}]{zeroshot}
Takeshi Kojima, Shixiang~Shane Gu, Machel Reid, Yutaka Matsuo, and Yusuke Iwasawa. 2022.
\newblock \href {https://dl.acm.org/doi/10.5555/3600270.3601883} {Large language models are zero-shot reasoners}.
\newblock \emph{Advances in neural information processing systems}, 35:22199--22213.

\bibitem[{Lappin(2024)}]{strongweak}
Shalom Lappin. 2024.
\newblock \href {https://link.springer.com/article/10.1007/s10849-023-09409-x} {Assessing the strengths and weaknesses of large language models}.
\newblock \emph{Journal of Logic, Language and Information}, 33(1):9--20.

\bibitem[{Manyika and Hsiao(2023)}]{bard}
James Manyika and Sissie Hsiao. 2023.
\newblock \href {https://ai.google/static/documents/google-about-bard.pdf} {An overview of bard: an early experiment with generative ai}.
\newblock \emph{AI. Google Static Documents}, 2.

\bibitem[{Min et~al.(2022)Min, Lyu, Holtzman, Artetxe, Lewis, Hajishirzi, and Zettlemoyer}]{icl2}
Sewon Min, Xinxi Lyu, Ari Holtzman, Mikel Artetxe, Mike Lewis, Hannaneh Hajishirzi, and Luke Zettlemoyer. 2022.
\newblock \href {https://doi.org/10.18653/v1/2022.emnlp-main.759} {Rethinking the role of demonstrations: What makes in-context learning work?}
\newblock In \emph{Proceedings of the 2022 Conference on Empirical Methods in Natural Language Processing}, pages 11048--11064, Abu Dhabi, United Arab Emirates. Association for Computational Linguistics.

\bibitem[{Nam et~al.(2023)Nam, Chung, and Hong}]{nam}
Yunju Nam, Hyenyeong Chung, and Upyong Hong. 2023.
\newblock \href {https://www.liebertpub.com/doi/abs/10.1089/cyber.2022.0356?journalCode=cyber} {Language artificial intelligences' communicative performance quantified through the gricean conversation theory}.
\newblock \emph{Cyberpsychology, Behavior, and Social Networking}, 26(12):919--923.

\bibitem[{Orr{\`u} et~al.(2023)Orr{\`u}, Piarulli, Conversano, and Gemignani}]{diffnlp1}
Graziella Orr{\`u}, Andrea Piarulli, Ciro Conversano, and Angelo Gemignani. 2023.
\newblock \href {https://www.frontiersin.org/journals/artificial-intelligence/articles/10.3389/frai.2023.1199350/full} {Human-like problem-solving abilities in large language models using chatgpt}.
\newblock \emph{Frontiers in artificial intelligence}, 6:1199350.

\bibitem[{Park et~al.(2024)Park, Kim, Kim, Cho, Kim, Lee, Kim, and Lee}]{open-ko-llm}
Chanjun Park, Hyeonwoo Kim, Dahyun Kim, SeongHwan Cho, Sanghoon Kim, Sukyung Lee, Yungi Kim, and Hwalsuk Lee. 2024.
\newblock \href {https://aclanthology.org/2024.acl-long.177} {Open {K}o-{LLM} leaderboard: Evaluating large language models in {K}orean with {K}o-h5 benchmark}.
\newblock In \emph{Proceedings of the 62nd Annual Meeting of the Association for Computational Linguistics (Volume 1: Long Papers)}, pages 3220--3234, Bangkok, Thailand. Association for Computational Linguistics.

\bibitem[{Satpute and Agrawal(2022)}]{nlpprag1}
R.S. Satpute and A.~Agrawal. 2022.
\newblock \href {https://fzgxjckxxb.com/wp-content/uploads/2022/12/54-JBS1740.pdf} {Pragmatic analysis in natural language processing}.
\newblock \emph{Journal of Basic Sciences}, 22(12).

\bibitem[{Seals and Shalin(2023{\natexlab{a}})}]{prag4ai2}
SM~Seals and Valerie~L Shalin. 2023{\natexlab{a}}.
\newblock \href {https://arxiv.org/abs/2304.14543} {Discourse over discourse: The need for an expanded pragmatic focus in conversational ai}.
\newblock \emph{arXiv preprint arXiv:2304.14543}.

\bibitem[{Seals and Shalin(2023{\natexlab{b}})}]{prag4ai}
SM~Seals and Valerie~L Shalin. 2023{\natexlab{b}}.
\newblock \href {https://arxiv.org/abs/2310.18435} {Expanding the set of pragmatic considerations in conversational ai}.
\newblock \emph{arXiv preprint arXiv:2310.18435}.

\bibitem[{Shidiq(2023)}]{diffnlp2}
Muhammad Shidiq. 2023.
\newblock \href {https://ejournal.unuja.ac.id/index.php/icesh/article/view/5614} {The use of artificial intelligence-based chat-gpt and its challenges for the world of education; from the viewpoint of the development of creative writing skills}.
\newblock In \emph{Proceeding of international conference on education, society and humanity}, volume~1, pages 353--357.

\bibitem[{Singh et~al.(2023)Singh, Cambronero, Gulwani, Le, Negreanu, and Verbruggen}]{parameters}
Mukul Singh, Jos{\'e} Cambronero, Sumit Gulwani, Vu~Le, Carina Negreanu, and Gust Verbruggen. 2023.
\newblock \href {https://doi.org/10.18653/v1/2023.emnlp-main.716} {{C}ode{F}usion: A pre-trained diffusion model for code generation}.
\newblock In \emph{Proceedings of the 2023 Conference on Empirical Methods in Natural Language Processing}, pages 11697--11708, Singapore. Association for Computational Linguistics.

\bibitem[{Son et~al.(2024)Son, Lee, Kim, Kim, Muennighoff, Choi, Park, Yoo, and Biderman}]{kmmlu}
Guijin Son, Hanwool Lee, Sungdong Kim, Seungone Kim, Niklas Muennighoff, Taekyoon Choi, Cheonbok Park, Kang~Min Yoo, and Stella Biderman. 2024.
\newblock \href {https://arxiv.org/abs/2402.11548} {Kmmlu: Measuring massive multitask language understanding in korean}.
\newblock \emph{arXiv preprint arXiv:2402.11548}.

\bibitem[{Sudirjo et~al.(2023)Sudirjo, Diantoro, Al-Gasawneh, Azzaakiyyah, and Ausat}]{easynlp1}
Frans Sudirjo, Karno Diantoro, Jassim~Ahmad Al-Gasawneh, Hizbul~Khootimah Azzaakiyyah, and Abu Muna~Almaududi Ausat. 2023.
\newblock \href {https://jurnal.unidha.ac.id/index.php/jteksis/article/view/871} {Application of chatgpt in improving customer sentiment analysis for businesses}.
\newblock \emph{Jurnal Teknologi Dan Sistem Informasi Bisnis}, 5(3):283--288.

\bibitem[{Team et~al.(2023)Team, Anil, Borgeaud, Wu, Alayrac, Yu, Soricut, Schalkwyk, Dai, Hauth et~al.}]{gemini}
Gemini Team, Rohan Anil, Sebastian Borgeaud, Yonghui Wu, Jean-Baptiste Alayrac, Jiahui Yu, Radu Soricut, Johan Schalkwyk, Andrew~M Dai, Anja Hauth, et~al. 2023.
\newblock \href {https://arxiv.org/abs/2312.11805} {Gemini: a family of highly capable multimodal models}.
\newblock \emph{arXiv preprint arXiv:2312.11805}.

\bibitem[{Touvron et~al.(2023)Touvron, Lavril, Izacard, Martinet, Lachaux, Lacroix, Rozi{\`e}re, Goyal, Hambro, Azhar et~al.}]{llama}
Hugo Touvron, Thibaut Lavril, Gautier Izacard, Xavier Martinet, Marie-Anne Lachaux, Timoth{\'e}e Lacroix, Baptiste Rozi{\`e}re, Naman Goyal, Eric Hambro, Faisal Azhar, et~al. 2023.
\newblock \href {https://arxiv.org/abs/2302.13971} {Llama: Open and efficient foundation language models}.
\newblock \emph{arXiv preprint arXiv:2302.13971}.

\bibitem[{Wei et~al.(2022)Wei, Wang, Schuurmans, Bosma, Xia, Chi, Le, Zhou et~al.}]{cot}
Jason Wei, Xuezhi Wang, Dale Schuurmans, Maarten Bosma, Fei Xia, Ed~Chi, Quoc~V Le, Denny Zhou, et~al. 2022.
\newblock \href {https://dl.acm.org/doi/10.5555/3600270.3602070} {Chain-of-thought prompting elicits reasoning in large language models}.
\newblock \emph{Advances in neural information processing systems}, 35:24824--24837.

\bibitem[{Yang et~al.(2024)Yang, Jin, Tang, Han, Feng, Jiang, Zhong, Yin, and Hu}]{llm1}
Jingfeng Yang, Hongye Jin, Ruixiang Tang, Xiaotian Han, Qizhang Feng, Haoming Jiang, Shaochen Zhong, Bing Yin, and Xia Hu. 2024.
\newblock \href {https://dl.acm.org/doi/10.1145/3649506} {Harnessing the power of llms in practice: A survey on chatgpt and beyond}.
\newblock \emph{ACM Transactions on Knowledge Discovery from Data}, 18(6):1--32.

\bibitem[{Yoo et~al.(2024)Yoo, Han, In, Jeon, Jeong, Kang, Kim, Kim, Kim, Kim et~al.}]{clova}
Kang~Min Yoo, Jaegeun Han, Sookyo In, Heewon Jeon, Jisu Jeong, Jaewook Kang, Hyunwook Kim, Kyung-Min Kim, Munhyong Kim, Sungju Kim, et~al. 2024.
\newblock \href {https://arxiv.org/abs/2404.01954} {Hyperclova x technical report}.
\newblock \emph{arXiv preprint arXiv:2404.01954}.

\bibitem[{Zhao et~al.(2023)Zhao, Zhou, Li, Tang, Wang, Hou, Min, Zhang, Zhang, Dong et~al.}]{llm2}
Wayne~Xin Zhao, Kun Zhou, Junyi Li, Tianyi Tang, Xiaolei Wang, Yupeng Hou, Yingqian Min, Beichen Zhang, Junjie Zhang, Zican Dong, et~al. 2023.
\newblock \href {https://arxiv.org/abs/2303.18223} {A survey of large language models}.
\newblock \emph{arXiv preprint arXiv:2303.18223}.

\end{thebibliography}
\bibliographystyle{acl_natbib}

\end{document}